\documentclass[9pt]{article}

\usepackage[margin=0.82in]{geometry}
\usepackage[utf8]{inputenc}
\usepackage[T1]{fontenc}
\usepackage{microtype}

\usepackage{graphicx}
\usepackage{subcaption}
\usepackage{overpic}
\usepackage{booktabs}

\usepackage{amsmath, amssymb, amsthm}
\usepackage{amsfonts}

\usepackage{xcolor}
\usepackage[textsize=tiny]{todonotes}

\usepackage{xurl}
\usepackage{url}
\usepackage[hidelinks]{hyperref}

\usepackage[square,sort]{natbib}
\let\cite\citep 

\usepackage[capitalize,noabbrev]{cleveref}

\DeclareMathOperator{\R}{\mathbb{R}}
\DeclareMathOperator{\N}{\mathbb{N}}
\DeclareMathOperator{\scale}{\text{sc}}

\DeclareMathOperator{\Ima}{\text{Im}}
\DeclareMathOperator{\sign}{\text{sign}}
\DeclareMathOperator{\Span}{\text{span}}

\theoremstyle{plain}

\theoremstyle{definition}

\theoremstyle{remark}

\usepackage{titlesec}
\titleformat{\section}  
  {\fontsize{10}{10}\bfseries} 
  {\thesection} 
  {.5em} 
  {} 
  [] 

\titleformat{\subsection}  
  {\fontsize{9}{9}\bfseries} 
  {\thesubsection} 
  {.5em} 
  {} 
  [] 

\titleformat{\paragraph}[runin]
  {\fontsize{9}{9}\bfseries}
  {\theparagraph}{}{}[]

\title{PolyNODE: Variable-dimension Neural ODEs on M-polyfolds}
\author{Per {\AA}hag \and Alexander Friedrich \and Fredrik Ohlsson \and Viktor Vigren N{\"a}slund}
\date{} 

\begin{document}

\twocolumn[
  \maketitle
  \begin{center}
    \small
    Department of Mathematics and Mathematical Statistics, Ume{\aa} University, Ume{\aa}, Sweden\\
    \texttt{per.ahag@umu.se, alexander.friedrich@umu.se, fredrik.ohlsson@umu.se, viktor.vigren.naslund@umu.se}\\
    Authors are listed alphabetically according to their surname.
  \end{center}

  \vspace{0.5em}
  \begin{abstract}
  Neural ordinary differential equations (NODEs) are geometric deep learning models based on dynamical
  systems and flows generated by vector fields on manifolds. Despite numerous successful applications,
  particularly within the flow matching paradigm, all existing NODE models are fundamentally constrained
  to fixed-dimensional dynamics by the intrinsic nature of the manifold's dimension. In this paper, we
  extend NODEs to M-polyfolds (spaces that can simultaneously accommodate varying dimensions and a
  notion of differentiability) and introduce PolyNODEs, the first variable-dimensional flow-based model
  in geometric deep learning. As an example application, we construct explicit M-polyfolds featuring
  dimensional bottlenecks and PolyNODE autoencoders based on parametrised vector fields that traverse
  these bottlenecks. We demonstrate experimentally that our PolyNODE models can be trained to solve
  reconstruction tasks in these spaces, and that latent representations of the input can be extracted and
  used to solve downstream classification tasks. The code used in our experiments is publicly available at
  \texttt{https://github.com/turbotage/PolyNODE}.
  \end{abstract}

  \vspace{2em}
]

\section{Introduction}

Many modern neural networks change the dimension of their hidden representation across layers.
They expand features to increase expressive power or compress them to extract information-dense latent representations.
The ability to accommodate variable feature space dimensions is central in encoder-decoder models, convolutional neural networks and many other ubiquitous feed-forward machine learning architectures.
A natural question is what happens if we take the continuous depth viewpoint in this setting?
What is a continuous depth model whose state dimension is allowed to change over time?

Neural ordinary differential equations (NODEs) provide the continuous depth limit of residual type networks with constant width layers by replacing discrete layers with the flow of an ordinary differential equation (ODE)~\cite{Chen2018}.
In its basic form, the hidden state evolves as
\begin{equation}
\label{eq:dyn-syst}
\frac{d}{dt}\phi(t,p) = X(\phi(t,p))\,, \qquad \phi(0,p) = p \,,
\end{equation}
where $X: \R^n \to \R^n$ is the vector field generating the flow $\phi: \R \times \R^n \to \R^n$ and $h(p)=\phi(1,p)$ is the diffeomorphism defining the map between the input and the output of the NODE model.
This construction extends to manifold valued states by replacing $\R^n$ with a manifold $M$ and generalising to flows $\phi : \mathbb{R} \times M \to M$ generated by vector fields $X$ on $M$ \cite{Falorsi2020,Lou2020,Mathieu2020}.
NODEs are widely used in flow based learning and generative modelling, and features attractive properties such as universality~\cite{Zhang2020,Andersdotter2024} and scalable training using flow matching~\cite{Liu2023,Lipman2023,Albergo2023,Tong2024}.

In \eqref{eq:dyn-syst}, the state $\phi(t,p)$ always lies in a space of fixed dimension, either $\R^n$ or a manifold $M$.
This is not a limitation of parametrisation but an intrinsic  geometric restriction.
A smooth manifold does not allow trajectories whose number of local degrees of freedom drop or increase at an intermediate time.
For this reason, NODEs do not provide an intrinsic continuous depth analogue of, e.g.~encoder-decoder architectures~\cite{LeCun1987,Bourlard1988,Baldi1989,Hinton1989} where a dimensional bottleneck  is used to extract a latent representation from which the original data can be efficiently reconstructed.
More generally, NODEs cannot capture variable width feed forward network dynamics in continuous time \cite{RuizBalet2023}.

If we insist on changing feature dimensions, then the correct state space is no longer a manifold.
The continuous time analogue is instead a stratified space, where smooth pieces of different dimensions meet along singular loci.
A simple example of such a space corresponding to the dimensional bottleneck of the autoencoder is $\Omega^m_n := \left( (-\infty, \tau_1) \times \R^{n} \times \R^m \right) \cup \left( [\tau_1, \tau_2] \times \R^{n} \times \{0\} \right) \cup \left((\tau_2, \infty) \times \R^{n} \times \R^m \right)$.
Outside $[\tau_1,\tau_2]$ trajectories evolve in $(\R \setminus [\tau_1,\tau_2] ) \times \R^n \times \R^m$, but on $[\tau_1,\tau_2]$ they are constrained to $[\tau_1,\tau_2] \times \R^n \times \{0\}$.
At $\tau_1$ and $\tau_2$ the local dimension changes, and at these these points there are no manifold charts since the space is not locally homeomorphic to $\mathbb{R}^n$. Consequently, the standard manifold calculus that underlies NODE theory breaks down.

A natural approach to extend NODEs would be to embed $\Omega^m_n$ into an ambient Euclidean space and define an ordinary NODE there.
However, this does not resolve the main issue which is the calculus at the singular set.
If we require the ambient vector field to be smooth and tangent to the stratified set, then its flow cannot probe a genuine bottleneck in finite time.
Finite time collapse of the transverse coordinates forces the dynamics to become singular or at least non-Lipschitz near the transition, and then the usual sensitivity calculus used for training NODEs is no longer justified.
What we need is an intrinsic notion of smoothness on the stratified state space that keeps a chain rule and a well defined tangent map while still allowing controlled collapse.

M-polyfold theory provides exactly this \cite{Hofer2007,Hofer2009a,Hofer2009b,Weber2019, Hofer2021,Aahag2025}.
It was developed to give a workable differential calculus on spaces that arise from gluing and degeneration, where effective dimension changes are part of the geometry.
In our setting, the key point is that an M-polyfold equips a stratified space with a global notion of differentiability that records how the collapsing coordinates vanish as one approaches the bottleneck.
This replaces the missing manifold structure at $\tau_1$ and $\tau_2$ and restores the tools needed for flow based learning, including a chain rule and a tangent construction that are compatible with the bottleneck.

In this paper we use the M-polyfold framework to extend NODEs beyond manifolds.
We introduce PolyNODEs, continuous depth models driven by parametrised vector fields on explicit M-polyfolds.
We explicitly construct M-polyfolds that realise dimensional bottlenecks and PolyNODE autoencoders that encode and decode by traversing these bottlenecks.
We then demonstrate empirically that these models can be trained for reconstruction tasks and that the resulting latent representations support downstream classification.

Our main contributions are as follows.
\begin{itemize}
\item We introduce PolyNODEs, a class of continuous depth models on M-polyfolds that allows variable dimension dynamics.
\item We construct explicit M-polyfold bottleneck spaces and PolyNODE autoencoders by para\-metrising vector fields whose flows traverse the bottlenecks.
\item We provide proof of concept experiments showing that PolyNODE autoencoders can be trained for reconstruction and that the learned latent codes support classification.
\end{itemize}

\section{Neural ODEs on M-polyfolds}

\begin{figure}[!ht]
    \centering
    \includegraphics[width=0.8\linewidth]{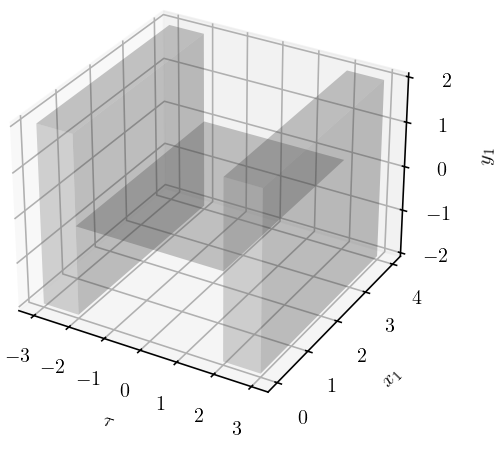}
    \caption{Illustration (grey shading) of the stratified topological space $\Omega^1_1 \subset \R^3$ with $\tau_1=-2$ and $\tau_2=2$.}
    \label{fig:Omega11}
\end{figure}

\begin{figure}[!ht]
    \centering
    \includegraphics[width=0.8\linewidth]{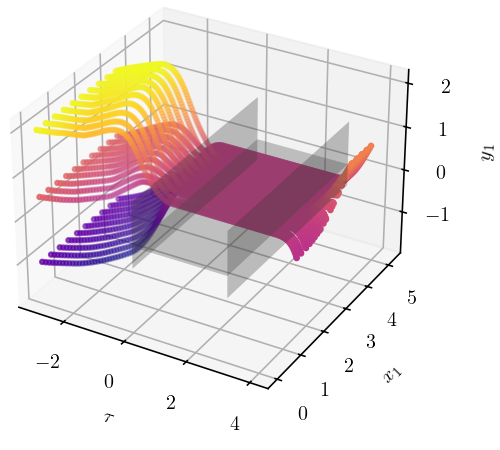}
    \caption{Illustration of a semi-flow in $\Omega^1_1 \subset \R^3$ with $\tau_1=-2$ and $\tau_2=2$, polyfold structure indicated as grey planes.}
    \label{fig:flow_illu}
\end{figure}

The description of the stratified spaces required to accommodate NODE state vectors that can change their dimension -- its regular components, their boundaries and intersections -- is quite cumbersome even for simple configurations like the space $\Omega^m_n$ introduced above and illustrated in Figure~\ref{fig:Omega11}.  
Generalising NODEs to stratified spaces, the regular components correspond to regions of fixed dimension dynamics, and the boundaries to discontinuous changes in the dimension of the state vector $\phi(t,p)$, as illustrated in Figure~\ref{fig:flow_illu}.
The concept of a M-polyfold allows a consistent description of stratified spaces like $\Omega^m_n$. More importantly, polyfold theory endows these spaces with a global smooth structure which gives us access to much the same analytical tools that are available on manifolds. In particular, we have well-defined notions of vector fields and can construct the flows required to extend NODEs to these spaces.
Below we present a very brief introduction to the theory of M-polyfolds based on~\cite{Hofer2021}. Subsequently, we give an explicit M-polyfold structure for the stratified space $\Omega^m_n$, construct flows and ODEs on M-polyfolds, and define the PolyNODE model which is our first main contribution.
We use $\Omega^m_n$ as a recurring example, as it features in our construction of PolyNODE autoencoders in Section~\ref{sec:NODE_autoencoder}, and denote a point in this space as $p=(\tau,x,y)$ with $\tau \in \mathbb{R}$, $x\in \mathbb{R}^n$, and $y \in \mathbb{R}^m$.

\subsection{Differential Geometry of M-polyfolds}
\label{subsec:DG_mpoly}

\paragraph{sc-Banach Spaces}
A \emph{scale Banach space}, or sc-Banach space, is a Banach space $E$, together with a filtration $\{E_k\}_{k\in \N}$ called a \emph{scale structure}, or sc-structure, where $E_0 =E$ and $E_{k+1} \subset E_k$ for all $k \in \N$, all inclusions $\iota_k: E_{k+1} \hookrightarrow E_k $ are compact, and the space $E_\infty := \bigcap_{k\in \N}E_k$ is dense in every \emph{scale} $E_k$.

Each $E_k$ can be turned into a sc-Banach space of its own, denoted by $E^k$, where $(E^k)_i := E_{k+i}$ for all $i\in \N$. Direct sums of sc-Banach spaces are again sc-Banach spaces with component-wise scales. Note that finite dimensional Banach spaces admit only the constant scale structure, i.e.~$E_k=E$ for all $k\in \N$, since the only dense vector subspace of $E$ is $E$ itself. Infinite dimensional Banach spaces on the other hand cannot be equipped with the constant scale structure.

In this article we are mainly interested in the case of $L^2(\R)$ with a weighted Sobolev sc-structure. Let $\{\delta_k\}_{k=0}^\infty$ be a non-negative, strictly increasing sequence with $\delta_0=0$, let $\alpha:\R \to [-1,1]$ be a smooth odd function with $\alpha(0)=0$ and $\alpha(s)=1$ for $s>1$.
The weighted Sobolev norms are then given by 
\begin{align*}
    \|f\|^2_k :=\int_{\R} \sum_{i=0}^k |d^i f(s)|^2 e^{\delta_k s \alpha(s)} \, ds\,, 
\end{align*}
and the scales of $L^2(\R)$ are defined by
\begin{align}
\label{eq:sobolev_scales}
    L^2(\R)_k := \big\{ f \in L^2(\R) \, \big| \, \|f\|_k < \infty \big\}\,.    
\end{align}
The inclusion $\iota : L^2(\R)_{k+1} \hookrightarrow L^2(\R)_k$ is compact by embedding theorems for weighted Sobolev spaces, compare Lemma 4.10 in~\cite{Fabert2016}, and $E_\infty$ contains $C_c^\infty(\R)$ which is dense in any Sobolev space.

In this example it is clear that points in higher scales have higher regularity and decay rates at infinity since the higher norms incorporate more derivatives and the exponential weights are increasing.

For two sc-Banach spaces $E$ and $F$ a map $\psi :E \to F$ is called \emph{scale continuous}, or $\scale^0$, if $\psi|_{E_k}: E_k \to F_k$ is continuous for all $k\in \mathbb{N}$. Furthermore, a sc-continuous map $\psi$ is called \emph{scale differentiable}, or $\scale^1$, if there exists a sc-continuous map $D\psi: E^1 \oplus E \to F$, such that $D\psi_{\xi}:E \to F$ is a bounded linear operator and  for $\xi, h \in E_1$
\[ \lim_{\|h\|_1 \to 0} \frac{\|\psi(\xi+h) - \psi(\xi) - D\psi_{\xi}(h) \|_0}{\|h\|_1}  = 0.\]
The tangent space of $E$ at a point $\xi \in E^1$ is defined as $T_{\xi} E:= E$.

Note that sc-continuity is a stronger condition than ordinary continuity. Sc-differentiability on the other hand is weaker than ordinary differentiability, assuming sc-continuity is established, since we only require $D\psi$ to exist on $E_1$, and use the stronger convergence for $h$ in the $E_1$ norm.

A map $\psi:E \to F$ is called $\scale^k$, for $k\in \N$, if the above construction can be iterated $k$ times for the respective differentials and \textit{scale smooth}, or $\scale^{\infty}$, if it is $\scale^k$ for any $k$.

\paragraph{Retracts}
In the theory of M-polyfolds, retracts play a similar role as open sets of $\R^n$ do for manifolds. A $\scale$-smooth \emph{retraction} on a sc-Banach space $E$ is a $\scale$-smooth map $r: E \to E$ that satisfies the projection property $r\circ r = r$. The image of a $\scale$-smooth retraction is called a $\scale$-smooth \emph{retract}.
If $r$ is a retraction and $O:=\Ima r$ the associated retract, the tangent space of $O$ at $\xi\in O \cap E_1$ is defined as $\Ima Dr_{\xi}$.

A \textit{manifold-like polyfold}, or M-polyfold, is a paracompact Hausdorff space that admits an atlas with charts constructed out of retracts, giving rise to a rich structure in the same way as on manifolds. For the purposes of this paper we are content with a single retract. This is analogous to a manifold that can be covered by a single chart.

\paragraph{$\Omega^1_1$ as an M-polyfold} We now return to the space $\Omega^1_1 $ and show that it can be described by a single retract. Let $\tau_1, \tau_2 \in \R$, with $\tau_1 < \tau_2$, set $J:= [ \tau_1, \tau_2]$, and define $\beta: \R\setminus J\to \R$ as
\begin{equation*}
    \beta(\tau) = \exp\left( 1/(\tau_1 - \tau) + 1/(\tau-\tau_2) \right)\,.
\end{equation*}
Let $L^2(\R)$ be the scale Banach space with weighted Sobolev scales in~\eqref{eq:sobolev_scales}, fix a $\gamma \in C_c^\infty(\R) \subset E_\infty$ with $\| \gamma\|_{L^2(\R)}=1$, define $\gamma_\tau(s) := \gamma(s + \beta(\tau))$, and abbreviate $\rho_\tau := d/d\tau\, \gamma_\tau$. We then define a $sc$-Banach space $E:= \R \oplus \R \oplus L^2(\R)$, denote a point in this space as $\xi = (\tau,x,f)$ with $\tau \in \mathbb{R}$, $x\in\mathbb{R}$, and $f \in L^2(\mathbb{R})$, and define the map $r:E \to E$,
\begin{align}
\label{eq:retract}
    r(\tau, x, f) =
    \begin{cases}
        (\tau, x, \langle f, \gamma_\tau \rangle \gamma_\tau),  & \tau\in \R \setminus J \\
        (\tau, x, 0),  & \tau\in J
    \end{cases}\,,
\end{align}
where $\langle \cdot,\cdot \rangle$ denotes the standard inner product on $L^2(\R)$. Explicit calculations show that $r$ is $\scale^\infty$, with differential
\begin{align*}
    Dr_{(\tau, x, f)}(\sigma, y, g) =
    \begin{cases}
        \left(\sigma, y,  \sigma \langle f, \gamma_\tau \rangle \rho_\tau \right. \\
        \quad + \sigma \langle f, \rho_\tau \rangle \gamma_\tau \\
        \left. \quad + \langle g, \gamma_\tau \rangle  \gamma_\tau \right) , & \tau\in \R \setminus J \\
        (\sigma, y, 0), & \tau\in J
    \end{cases}\,,
\end{align*}
and it clearly satisfies the projection property making it a retraction on $E$.

In the above construction the function $\beta$ has singularities at $\tau_1 $ and $\tau_2$. This means the support of $\gamma_\tau$ is shifted to infinity at these points. This in turn yields strong decay estimates for the term $\langle f, \gamma_\tau \rangle$ on every scale, which leads to the apparent jump in \eqref{eq:retract} not only being continuous but sc-smooth. 
We can think of the curves $\tau \mapsto r(\tau,f)$ decaying to $0$ in the last component as $\tau \nearrow \tau_1$ and $\tau \searrow \tau_2$ for $f$ in any scale.

The retract $O = \Ima r$ is homeomorphic to $\Omega^1_1 = \left( \R \setminus J \times \R \times \R \right) \cup \left( J \times \R \times \{0\} \right)$, with the subspace topology of $\R^3$, via the map $\eta: O \to \Omega^1_1$
\begin{align}
\label{eq:eta_chart}
    \eta(\tau, x,  f) = 
    \begin{cases}
        (\tau, x,  \langle f , \gamma_\tau \rangle ), & \tau\in \R \setminus J \\
        (\tau, x, 0), & \tau \in J
    \end{cases}\,.
\end{align}
At $\tau\in \R \setminus J$, the tangent space of $O$ is  \begin{equation*}
    T_{(\tau, x, f)}O = \Span \left\{ \left(  1, 0,  \langle f, \gamma_\tau \rangle \rho_\tau \right), (0,1,0), (0, 0, \gamma_\tau)\right\} ,
\end{equation*}
whereas the tangent space of $\Omega^1_1$ is simply $\R^3$,
\begin{align*}
    T_{\eta(\tau, x, f)} \Omega^1_1 & = D\eta_{(\tau, x, f)} (T_{(\tau, x, f)}O) \\ & = \Span \left\{ (1,0,0), (0,1,0), (0,0,1) \right\}\,,
\end{align*}
due to the fact that $\gamma_\tau$ and $\rho_\tau$ are $L^2(\R)$-orthogonal\footnote{Consider the $\tau$ derivative of $1= \|\gamma_\tau\|^2_{L^2(\R)}$.}. 

The constructions of this section generalise in a straightforward manner. The second coordinate, denoted by $x\in \R$ above, can be replaced by a coordinate in $\R^n$ since the construction does not depend on it. More importantly, we may increase the jump in dimensions to $m$ by choosing an $L^2$-orthonormal family $\{\gamma_i\}_{i=1}^m$ of smooth compactly supported functions instead of a single $\gamma$. The retraction $r$ is then given by $r(\tau,x,f) = (\tau,x, \pi_{\tau}(f))$, for $\tau\in \R \setminus J$, where $\pi_\tau$ is the $L^2$-projection to $\Span\{(\gamma_i)_\tau\}_{i=1}^m$. In this way we can construct M-polyfolds $\Omega^m_n \cong \left( \R \setminus J \times \R^n \times \R^m \right) \cup \left( J \times \R^n \times \{0\} \right) $ for arbitrary $n,\, m \in \mathbb{N}$.

\subsection{Flows and ODEs on M-polyfolds}
\label{subsec:flows_mpoly}

In order to extend NODE models to M-polyfolds we need vector fields compatible with the sc-smooth structure. A $\scale^k$ vector field on a sc-Banach space $E$ is a $\scale^k$ map $X: E^1 \to E$, and we denote the space of such vector fields on $E$ by $\mathfrak{X}^k(E)$. On a retract $O\subset E$, with retraction $r$, we can view the set of vector fields as $\mathfrak{X}^k(O) = \{X \in \mathfrak{X}^k(E) \mid X_{\xi} \in T_{\xi}O, \, \forall \xi\in O \cap E^1 \}$.

To construct the flow generated by a vector field $X$ we need to solve ODEs on $E$. If a vector field $X$ is actually in $\scale^0(E^1,E^1)$ then, since $E$ is a filtration of Banach spaces, ODE solution theory applies independently on every scale $E_k$. Thus ODEs on $E$ can be solved on every scale. If additionally $X$ is Lipschitz continuous on every scale then standard uniqueness arguments imply the solutions for different scales agree when restricted to the same scale. This generalises to the theory of flows by similar arguments for the dependence on the initial conditions. We discuss ODEs and flows on sc-Bannach spaces and M-polyfolds in greater detail in a forthcoming article. For ODEs on Banach spaces see for instance Chapter~16 in \cite{Pata2019}.

Given  $X \in \mathfrak{X}^k(O)$ on a retract $O$ with retraction $r$, we can consider a lift, i.e.~$\tilde{X} \in \mathfrak{X}^k(E)$ such that $ Dr_{\xi}\tilde{X}(\xi) = X(r(\xi))$. If $\tilde{X}$ can be integrated to a flow $\tilde{\phi}$ on $E^1$, then it gives rise to a flow $r\circ \tilde{\phi}$ on $O\cap E^1$ when restricted to $O\cap E^1$.

However, in the case of the retraction~\eqref{eq:retract} for the  $\Omega^1_1$ model this construction is not applicable, since at $\tau_1$ and $\tau_2$ any lift of a vector field whose flow traverses the bottlenecks will be singular as a map from $E^1$ to $E^1$.
This is expected because at these points any flow going into the dimensional bottleneck has to loose injectivity and thus ceases to be a flow.
This is a fundamental property of stratified spaces like $\Omega^1_1$ and indeed the main feature we use for the autoencoder construction.
Thus, what we are really interested in for the autoencoder application are semi-flows that traverse the dimensional bottleneck.
A semi-flow $\phi$ that is $\scale^0$ at $\tau_1$ and $\tau_2$  has a component $\phi_3$ which converges to $0$ at these points in any scale by definition of scale continuity.
In coordinate charts $\eta$~\eqref{eq:eta_chart} this corresponds to super-exponential decay.
For example, suppose $\lim_{t\to t_1} \phi(t)_1 = \tau_1$ from below, then 
\begin{align}
\label{eq:compressing_decay}
    \lim_{t\to t_1} \eta(\phi(t))_3 \exp \left( \delta_i \beta(\phi(t)_1)\right) =0 \quad \forall i\in \N.
\end{align}

In general, vector fields on the retract $O = \Ima r$ have the form
\[X_{(\tau,x, \langle f , \gamma_\tau \rangle \gamma_\tau)} = \left( X_1, X_2, X_1 \langle f , \gamma_\tau \rangle \rho_\tau + X_3 \gamma_\tau \right)\,,\]
for functions $X_i:O \to \R$, $i\in \{1,2,3\}$. This means that for $\tau\in \R\setminus J$ we can view vector fields on $\Omega^1_1 \subset \R^{3}$ as classical vector fields $Y_{(\tau, x, y)}=(Y_1,Y_2,Y_3)$, which lift to $\scale^0$ vector fields
\begin{align}
    \label{eq:lifted_vf}
    X_{(\tau,x, y \gamma_\tau)} &= D\eta^{-1}(Y (\eta(\tau, x, y \gamma_\tau))) \nonumber\\
    &= \left( Y_1, Y_2, Y_1 y \rho_\tau + Y_3 \gamma_\tau \right)\,.
\end{align}
Using techniques like separation of variables it is straight forward to find vector fields $Y = (Y_1, Y_2, Y_3)$ whose  semi-flows satisfy~\eqref{eq:compressing_decay}. Note  that condition~\eqref{eq:compressing_decay} is fulfilled if there is a $t_0<t_1$ such that $\eta(\phi(t))_2 =0$ for all $t\in [t_0, t_1]$. In Section~\ref{subsec:model_construction} we give an example of a vector field with an associated semi-flow with this property.

With slight abuse of notation, we let $\phi$ denote both the semi-flow on $\Omega^1_1$ generated by the vector field $Y$ and its image under $\eta^{-1}$.
See Figure~\ref{fig:flow_illu} for an illustration of a semi-flow in $\Omega^1_1$. The generalization to vector fields and semi-flows on $\Omega^m_n$ is straightforward, as indicated in Section~\ref{subsec:DG_mpoly}.

\subsection{PolyNODEs: Neural ODEs on M-polyfolds}
\label{subsec:scNODEs_mpoly}
Having defined the necessary machinery from polyfold theory, we can now extend the definition of a neural ODE to stratified spaces. Let $E$ be a sc-Banach space, $r:E \to E$ a retraction, and $O = \Ima r$ the corresponding retract. A \textit{PolyNODE} parametrised by the vector field $X \in \mathfrak{X}^k(O)$ is the map $h: O \to O$, $h(\xi) = \phi(1,\xi)$, where $\phi:\R \times O \to O$ is the semi-flow generated by $X$,
\begin{equation}
\label{eq:scNODE}
    \frac{d}{dt}\phi(t,\xi) = X(\phi(t,\xi))\,, \quad \phi(0,\xi) = \xi \,.
\end{equation}
The generalisation to M-polyfolds covered by several retracts is straightforward, but not required for the case $\Omega^m_n$ we consider below.

The semi-flow $\phi$ generated by $X$ can traverse the stratification points at the intersection of the regular components of the underlying space, meaning that the dimension of the state vector $\phi(t,\xi)$ can change in a $\scale$-differentiable way. This construction of PolyNODEs is our first major contribution; a flow-based machine learning model capable of accommodating variable-dimension dynamics.

\section{A PolyNODE Autoencoder Model on $\Omega^m_n$}
\label{sec:NODE_autoencoder}

Having defined PolyNODEs in the previous section, we proceed to construct explicit examples of such models. The purpose is to demonstrate the viability of PolyNODEs in machine learning by establishing; 1) that M-polyfolds corresponding to specific machine learning problems can be constructed; and 2) that these spaces admit vector fields which probe their M-polyfold structure. In the subsequent section we then present numerical experiments that further demonstrate; 3) that $sc$-NODE models defined by these vector fields can be trained to solve common types of machine learning tasks.

\subsection{Model Construction}
\label{subsec:model_construction}
\begin{figure}[ht]
    \centering
        \includegraphics[width=0.8\linewidth]{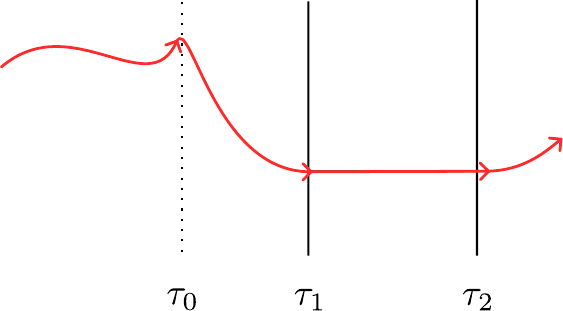}
    \caption{Illustration of a flow line entering the bottleneck at $\tau_1$ and exiting at $\tau_2$. Explicit compression starts at $\tau_0$.  }\label{fig:polyfold_tau_illustration}
\end{figure}

As a proof-of-concept, we consider the problem of reconstructing geometric objects using a PolyNODE with a dimensional bottleneck similar to that of a traditional autoencoder, for which the appropriate geometry is the space $\Omega^m_n$ with $J=[\tau_1,\tau_2]$ described in detail above. We can then interpret the region $\tau<\tau_1$ as the input space, the stratification at $\tau=\tau_1$ as the encoding, $\tau \in (\tau_1,\tau_2)$ as the latent space, the stratification at $\tau=\tau_2$ as decoding, and $\tau>\tau_2$ as the output space of the autoencoder structure, see Figure~\ref{fig:polyfold_tau_illustration} for an illustration.

For an index set $I$, we sample a submanifold $S \subset \R^{1+n+m}$ of dimension $\dim S \leq n+1$, $\{z_i\}_{i \in I}$, $z_i \in S$. We choose two embeddings $\iota_1$ for the input data and $\iota_2$ for the target data into $\Omega^m_n$, such that $\iota_1(z_i)_1 < \tau_1$ and $\iota_2(z_i)_1 > \tau_2$ for all $i\in I$.
An easy way to achieve this is to embed the samples $z_i \in S \subset \R^{n+m}$ into $\Omega^m_n$ as $\mathcal{X} = \{(\tau_{\mathcal{X}},z_i)\}_{i \in I}$ for a fixed $\tau_{\mathcal{X}}<\tau_1$ to obtain the input data and $\mathcal{Y} = \{(\tau_{\mathcal{Y}},z_i)\}_{i \in I}$ for some fixed $\tau_{\mathcal{Y}} > \tau_2$ to obtain the target data. We use this type of embedding in our main experiment, the autoencoding of the spiral, and assume it in the following.

For the reconstruction task, we restrict ourselves to vector fields whose semi-flows actually traverse the bottleneck. Furthermore, we regard the $\tau$ coordinate in $\Omega^m_n$ as artificial, corresponding to the depth in a traditional autoencoder, and consequently prescribe a constant velocity in this direction by enforcing $X_1= \tau_{\mathcal{Y}}-\tau_{\mathcal{X}}$.
A latent representation of a point $p \in \mathcal{X}$ is given by $\phi(t,p)$ for any $t$ such that $\phi(t,p)_1 \in [\tau_1, \tau_2]$. We may always choose $\Omega^m_n$ and the embeddings of the data such that this is realised at $t=1/2$.

\subsection{Parameterisation of Compressing Vector Fields}

In order to construct vector fields $X$ for the PolyNODE model that can accomplish the dimensional reduction in the encoding phase, we need to ensure convergence of the semi-flow to the stratification point $\tau=\tau_1$ in finite time. This is accomplished by choosing the parametrisation in~\eqref{eq:lifted_vf} and imposing conditions on the components $Y_{1+n+j}$, $j=1,\ldots,m$ compressing the semi-flow in the directions transverse to the latent space.

Let $\tau_0 < \tau_1$ define the compression region, let $k_j \in C^{0,1}(\R^{1+n+m}, \R_+)$, with $0<K_j<k_j$ for constants $K_j$, and let $a\in(0,1)$. For $p \in (\tau_0, \tau_1) \times \R^{n+m}$ we then restrict to vector fields
\begin{align}
    \label{eq:comp_vf}
    Y_{1+n+j}(p) &= -k_j(p) \sign(y_j) |y_j|^a \,,
\end{align}
which guarantees convergence of the semi-flow to $y_j=0$ in finite time~\cite{Bhat2000}. Specifically, for a constant rate function $k_j(p) = K_j$ the time to convergence is given by $T(y_j^0) = |y_j^0|^{1-a}/(K_j(1-a))$, where $y_j^0 \neq 0$ is the initial value. This means, by choosing the lower rate bounds $K_i$ appropriately, we can guarantee that the semi-flow $\phi$ maps a bounded subset of the region $ [\tau_0, \tau_1)\times \R^{n+m}$, containing the slice at $\tau=\tau_0$ of all flow lines originating on the input data $\mathcal{X}$, to $(\tau_0, \tau_1) \times \R^n \times \{0\}$ before it arrives at the stratification $\tau=\tau_1$. Thus $\phi$ restricted to that bounded set satisfies condition~\eqref{eq:compressing_decay} and is therefore an $\scale^0$ semi-flow.

We then obtain a family of compressing vector fields by parametrising the functions $k_j$ using neural networks. The components $Y_{1+i}$, $i=1,\ldots,n$ corresponding to the $x_i$ directions parallel to latent space are unrestricted and parametrised directly by neural networks. The same is true for $Y_{1+n+j}$ for $\tau<\tau_0$ and $\tau>\tau_2$.

Note that away from $y_j=0$, the $Y_{1+n+j}$ in~\eqref{eq:comp_vf} are smooth but at $y_j=0$ they are only Hölder continuous. ODE theory still guarantees the existence of a semi-flow but flow lines are no longer unique after reaching $y_j=0$, in particular flow lines may merge. When employing gradient based methods to train a PolyNODE with these compressing vector fields, the singularity at $y_j=0$ needs to be handled explicitly. See Appendix~\ref{app:backprop} for details of our implementation.

\section{Experiments}
\label{sec:experiments}
In this section we conduct several numerical experiments based on the PolyNODE autoencoder model constructed above\footnote{In this section, we use the term flow to refer to all collections of integer curves that appear, dropping the explicit semi-flow qualification.}. All experiments are based on parametrised families of the compressing vector fields in~\eqref{eq:comp_vf}. Complete experimental details are provided in Appendix~\ref{app:ex_details} the implementation is found in \cite{PolynodeCode2026}. In all experiments described below, we train our PolyNODEs using the adjoint sensitivity method~\cite{Chen2018} with a custom implementation of the backward pass for the compressing vector fields (see Appendix~\ref{app:backprop} for further details).

\subsection{Reconstruction of Geometrical Objects} \label{subsec:reconstruction}

\paragraph{Spirals}
\label{para:spiral_reconstr}
Our goal is to autoencode a family of spirals, where the number of turns ranges from $N=1/2$ to $N=5$, and extract a latent representation. 
To this end  we choose $\Omega^2_1$ with $\tau_1=0$ and $\tau_2=1$ as a M-polyfold, sample the spiral and embed to obtain the input data set $\mathcal{X} = \{p_i\}_{i\in I} = \{(\tau_\mathcal{X}, z_i)\}_{i\in I}$. The target data set is given by $\mathcal{Y} = \{(\tau_\mathcal{Y}, z_i)\}_{i\in I}$.
For the training of the model we employ a loss function $L = L_1 + \lambda(L_2 + L_3)$ with three loss terms, with weight $\lambda=20$ for the last two, where
\begingroup
\allowdisplaybreaks
\begin{align*}
L_1 &= \frac{1}{|I|} \sum\limits_{i\in I} |(  \phi(1,p_i) - (\tau_\mathcal{Y}, z_i) )|^2 \,, \\
L_2 &= \frac{1}{|I|} \sum\limits_{i\in I}   |\left(\phi(t_0,p_i)_3, \phi(t_0,p_i)_4 \right) - (1,1)|^2 \,, \\
L_3 &=  \frac{1}{|I|} \sum\limits_{i\in I} \Big|  d_S(p_i, q_{p_i}) - c_d|\phi(t_0,p_i)_2 - \phi(t_0,q_{p_i})_2| \Big| \,.
\end{align*}
\endgroup
Here  $c_d$ is the square root of the intrinsic diameter of the spiral; chosen this way since the intrinsic distance scales with the square of the angular parameter.Additionally, $q_p$ is a random element of $\mathcal{X}$, such that $q_p \neq p$, $\forall p \in \mathcal{X}$ and $q_p \neq q_{p'}$ if $p\neq p'$.

The term $L_1$ is the mean squared error of the reconstructed spiral.
The second term $L_2$ forces the flow to explore the fourth dimension and ensures that the $y_1$ and $y_2$ coordinates are in a range to be mapped to $0$ by the compressing vector fields by the time the flow reaches $\tau=0$.
The last term $L_3$ enforces the flow to be approximately isometric at time $t_0 = 1/4$, which aids in the unwinding of the spiral to a line.
Here $\tau_\mathcal{X}$ and $\tau_\mathcal{Y}$ are chosen such that at time $t_0$ the flow is at $\tau=\tau_0$, i.e.~the onset of the compressing vector fields.
For the spiral we know the distance function $d_S$ explicitly.
In general the approximate distance function may be reconstructed from the data, see for instance \cite{Memoli2005}. 
This last loss term is rather strong and problem specific, however it does not contain the entire information of the spiral parametrisation.
Moreover, we stress that the goal of this example is to show that a vector field for an  unwinding and autoencoding flow can be learned. 
In the next example, we show that a round sphere can be autoencoded using only the mean squared reconstruction error.
\begin{figure}[ht]
    \centering
    \begin{subfigure}[b]{0.49\linewidth}
        \begin{overpic}[width=\textwidth]{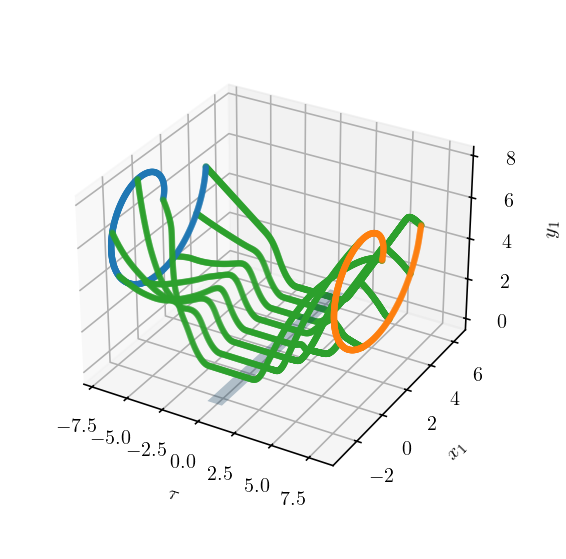}%
        \put(10,75){(a)}%
        \end{overpic}
    \end{subfigure}
    \begin{subfigure}[b]{0.49\linewidth}
        \begin{overpic}[width=\textwidth]{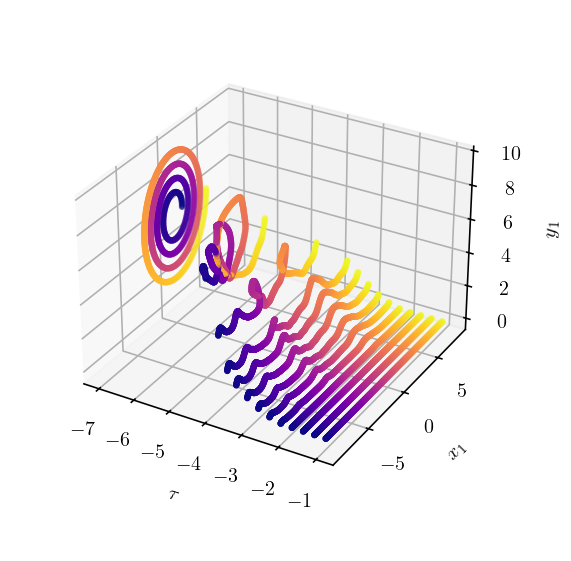}%
        \put(10,75){(b)}%
        \end{overpic}
    \end{subfigure}
    \caption{(a) Flow lines (green) for individual input samples for $N=1$. Input set $\mathcal{X}$ (blue) and reconstructed output $\hat{\mathcal{Y}}$ (orange). The $y_2$ component is projected out for visualisation.
    (b) Time slices of the flow for a spiral with $N=4$. Colour scale corresponds to the angular parameter.} \label{fig:spiral_flow}
\end{figure}
\begin{figure}[ht]
    \centering
        \includegraphics[width=0.95\linewidth]{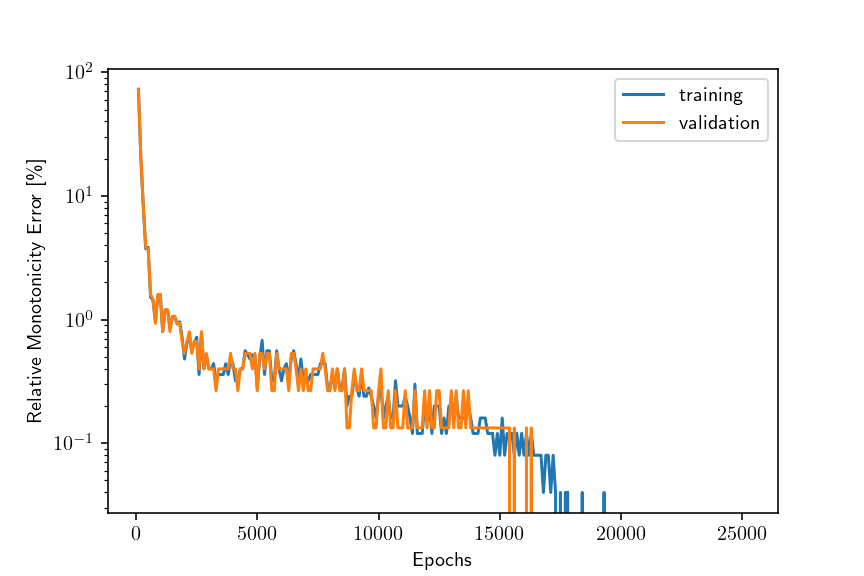}
    \caption{Relative monotonicity error during training for $N=4$.}\label{fig:monotonicity_training}
\end{figure}
Figure~\ref{fig:spiral_flow} illustrates the unwinding and encoding of the spirals to the line as well as the subsequent reconstruction. Note that the flow for $\tau < \tau_1$ is $4$ dimensional, thus allowing for apparent crossing of flow lines. 
To determine how well the spirals are encoded onto the line we may simply investigate the monotonicity of the map $s\mapsto \phi(1/2, f(s))_2$, where $f$ is the parametrisation of the spiral and $s\in[0,2\pi]$. Figure~\ref{fig:monotonicity_training} shows the monotonicity error relative to the number of sample points during training for a spiral with $N=4$. At the start of the training more than $40\%$ of the points are misaligned but as the training progresses the number of misaligned points goes to zero; shown by omission of points in the log scaled plot. This indicates that in this case the spiral is encoded perfectly onto the line, within the sample accuracy of $5000$ equidistant points. 

All of the experiments show very good monotonicity on sampled data points, with less than $0.02\%$ misaligned points. See Section~\ref{app:ex_spiral} for plots of the other experiments, including the reconstruction.

\paragraph{Sphere}
\label{para:sphere_reconstr}
As a further example, consider the two dimensional round sphere of radius $1$ in $\R^4$ around a point $x_0$ given by $ \{ (z_1, z_2, 0, z_3) \mid z\in S^2_1(x_0)\subset \R^3 \} $. We choose $\Omega^1_2$ with $\tau_1=0$ and $\tau_2=3$ as a M-polyfold, sample it and, with $e_1=(1,0,0,0)$, embedded the samples as  $ \mathcal{X}= \{p_i\}_{i\in I} = \{z_i+ \tau_\mathcal{X} e_1 \}_{i\in I} $ and $ \mathcal{Y} = \{z_i+ \tau_\mathcal{Y} e_1\}_{i\in I} $. Here we  break the embedding convention of Section~\ref{subsec:model_construction} for an easier visualisation.
The loss function is the  mean squared reconstruction error,
\[L = \frac{1}{I} \sum\limits_{i\in I} |(  \phi(1,p_i) - (z_i + \tau_\mathcal{Y}e_1   ))|^2 \,. \]

\begin{figure}[ht]
    \centering
    \begin{subfigure}[b]{0.49\linewidth}
        \begin{overpic}[width=\textwidth]{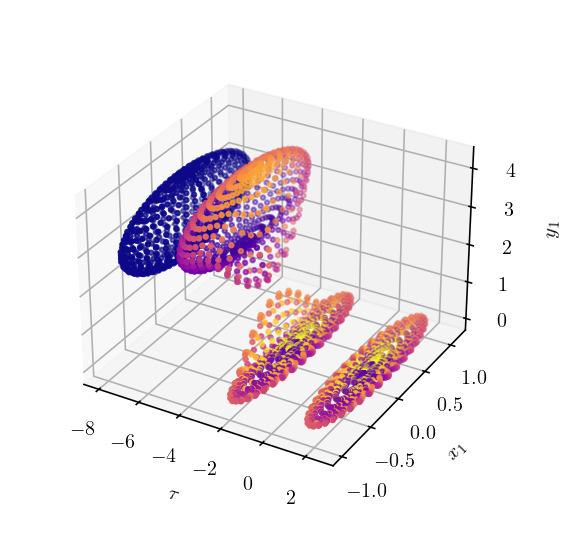}
        \put(10,75){(a)}
        \end{overpic}
    \end{subfigure}
    \begin{subfigure}[b]{0.49\linewidth}
        \begin{overpic}[width=\textwidth]{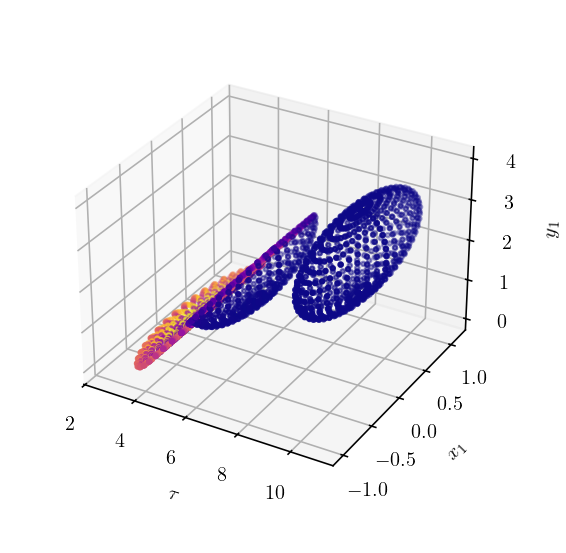}
        \put(10,75){(b)}
        \end{overpic}
    \end{subfigure}
    \caption{Time slices of the encoding (a) and decoding (b) of the $2$-sphere in $\R^4$. The colour represents the coordinate in the fourth dimension.}\label{fig:snapshots_sphere}
\end{figure}

Figure~\ref{fig:snapshots_sphere} shows the encoding and decoding of the $2$-sphere embedded in $\R^4$, where the fourth dimension is indicated by colour. We see that, starting from the monochrome sphere where the fourth coordinate is zero, the flow explores the extra dimension while the height is compressed. At the end of the encoding the sphere is contained in $\R^3$. At $0.06\%$, the relative mean reconstruction error of the decoded sphere is very low.

\subsection{Latent Space Classification Tasks}

Encoding–decoding architectures are often employed to enable downstream tasks based on the latent representation.
We illustrate this capability using the latent representation learned in the spiral experiment, Section~\ref{subsec:reconstruction}, to solve radial and angular classification tasks. In both cases, we define three labels $\{l_1, l_2, l_3\}$, described in detail in the following paragraphs, and extract latent states at $t=1/2$ to produce labelled samples for the classifications.

In principle, any model can be used for the classification task, but we opt to stay in the NODE framework. We designate target points in latent space corresponding to each label and learn a vector field whose flow maps the sample points to their corresponding label target.

The latent representation of the spiral is a one-dimen\-sional curve lying in a two-dimensional plane. For a general ordering of labels along the latent curve, we cannot guarantee the existence of a vector field making points flow to their target point if the integral curves are restricted to lie in the plane. This is because crossings of integral curves would be necessary for all but the simplest monotone orderings. Thus, we augment the NODE to three dimensions.

The loss used is MSE between the last points on the integral curves and their corresponding target points. In addition, an attractor term for each target point is added to the vector field to promote flow towards a target point, giving the NODE
\begin{equation}\label{eq:classification_vf}
    \frac{dx(t)}{dt} = Y_\theta(x,t) + C\sum_i\frac{y_i-x(t)}{|y_i-x(t)|}e^{-k(y_i-x(t))^2},
\end{equation}
where $y_i$ indicates the target point for label $l_i$ and $Y_\theta(x,t)$ is 
the learnable vector field.
We choose $k$ so that there is a clear 
separation between the target point attractors. 
For details on the
experimental setup and the neural networks used see Section~\ref{app:ex_classification}.

\paragraph{Radial Classification} We divide the interval~$\left[0, R_{\max}\right)$ into three equal parts, where $R_{max}$ is the maximal radius of the spiral. Each point in the latent space corresponds to a point on the spiral. A point in the latent space is given the label $l_i$, $i\in \{ 1,2,3\}$ if its corresponding point on the spiral has radius in $\left[R_{max}(i-1)/3,R_{max}i/3\right)$.

\begin{figure}[ht]
    \centering
    \begin{subfigure}[b]{0.49\linewidth}
        \centering
        \begin{overpic}[width=0.8\textwidth]
        {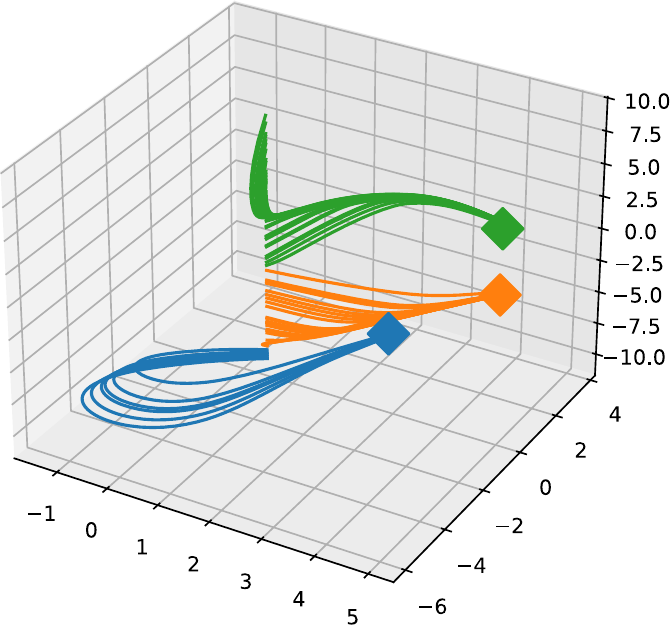}
        \put(0,80){(a)}
        \end{overpic}
    \end{subfigure}
    \begin{subfigure}[b]{0.49\linewidth}
        \centering
        \begin{overpic}[width=0.8\textwidth]{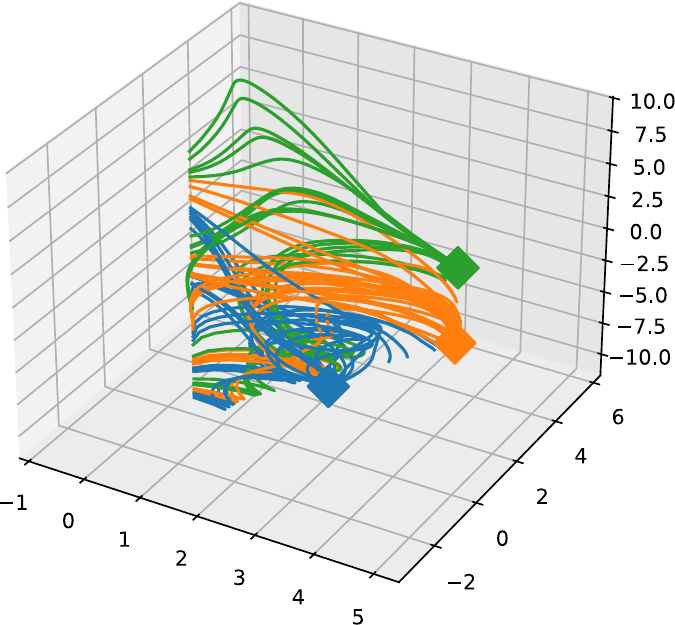}
        \put(0,80){(b)}
        \end{overpic}
    \end{subfigure}
    \caption{Classifier flow lines of the radial (a) and angular (b) classification problems. Colours indicate which label the trajectory belongs to and what label the target points (squares) 
    correspond to. The latent representations constituting the classifier input correspond to the line where the integral curves start.}\label{fig:spiral_flow_class}
\end{figure}

Due to the unwinding nature of the flow for the spiral auto-encoder, the labelling in latent space becomes monotone for the radial classification problem. Consequently, a simple flow is expected and experimentally observed, see Figure~\ref{fig:spiral_flow_class}a. 
The model reaches a peak accuracy of $100\%$ in this task, and consistently stays above $80\%$ accuracy after a single epoch.

\paragraph{Angular Classification} Classes are set by the angle in the plane of the spiral. The interval $\left[0,2\pi\right)$ is equally divided into 
$\left[2\pi(i-1)/3,2\pi i/3\right)$, $i \in \{1,2,3\}$ and a latent point is assigned a label $l_i$ according to its corresponding point on the spiral.

For angular classification, the unwinding nature of the autoencoder flow causes the labels to switch several times along the latent line. Therefore, no simple classification flow is possible and the augmentation is necessary. The experiment is qualitatively consistent with this expectation, see Figure~\ref{fig:spiral_flow_class}b.
The model reaches a peak accuracy of $98\%$ in this task 
and consistently exceeds $80\%$ accuracy after $150$ epochs.

With sufficiently many parameters and an expressive enough architecture, any desirable accuracy is achievable for both classification problems. Therefore, the accuracy presented should not be interpreted as an indication of superior model structure. 
Rather it demonstrates the ability of PolyNODEs to extract latent representations of sufficient accuracy for downstream tasks.

\section{Conclusion}
\label{sec:conclusion}

In this work, we have demonstrated that it is possible to extend flow-based NODE models to M-polyfolds to accommodate variable-dimension dynamics. The construction hinges on the use of infinite dimensional Banach spaces as the underlying spaces where the flow dynamics plays out, even in the case where the tangent spaces of the retracts and polyfolds we consider are finite dimensional. The reason is the fact that the non-trivial scale structures required to relax the differentiability condition and accomplish dimensional jumps cannot be defined in Euclidean spaces. 

We believe there are several interesting consequences of the construction of our PolyNODE models. The generality and flexibility of NODE models are increased by allowing varying dimension. We use the encoder-decoder structure as an example, but more generally PolyNODEs can be defined to represent the  continuum limit of feed forward networks with arbitrary layer widths. Similar architectural flexibility is often used in practice to design accurate and efficient neural networks. Somewhat conversely, the PolyNODEs could also be leveraged to gain a better understanding of well-posedness, stability and convergence for general architectures using ODE theory (cf.~\cite{Haber2018,Thorpe2023}).

We construct a M-polyfold mirroring an autoencoder setup, together with a family of parametrised vector fields that traverse the dimensional bottleneck. Our experiments with this class of models demonstrate the ability to train PolyNODE models and extract meaningful latent representations in an autoencoder setting, but contain no quantitative evaluation on more realistic problems or comparisons with other model architectures. Nor do we explore the vast range of possible geometries and architectures that could be constructed from flows on M-polyfolds. Both these aspects indicate interesting future directions of research.

Unlike the situation for Euclidean spaces (or even manifolds), there is currently no general theory for ODEs on M-polyfolds, meaning that the theoretical foundations for our PolyNODEs are not yet established.  Furthermore, an important question is whether the construction of PolyNODEs enables us to learn dynamics -- relevant for machine learning -- which cannot be described in terms of flows in the ambient space where finite-dimensional polyfolds can be embedded. For example, the stratified spaces we consider in this paper can always be embedded in Euclidean space $\Omega^m_n \subset \mathbb{R}^{1+n+m}$.  

In a forthcoming publication, we address both of these concerns by developing a theory of (semi-)flows on M-polyfolds generated by sc-smooth vector fields, and constructing examples of flows that traverse the bottleneck in $\Omega^m_n$ but cannot be extended to ambient space. We hope that this will establish the foundation for a new exciting research direction in variable-dimension flow-based machine learning.


 \section*{Acknowledgements}
The authors gratefully thank Aron Persson for many fruitful and insightful discussions.

The work of P{\AA} and FO was partially funded by the Swedish Research Council under grant agreement no.~2025-05053. The work of P{\AA}, FO, and VVN was partially supported by the Wallenberg AI, Autonomous Systems and Software Program (WASP) funded by the Knut and Alice Wallenberg Foundation. The work of AF was supported by a postdoctoral fellowship funded by Kempestiftelserna under grant number JCSMK24-0043.

The computations were enabled by resources provided by the National Academic Infrastructure for Supercomputing in Sweden (NAISS), partially funded by the Swedish Research Council through grant agreement no.~2022-06725.

\bibliographystyle{plainnat}
\bibliography{PolyNODE}

@book {Weber2019,
    AUTHOR = {Weber, Joa},
     TITLE = {Scale calculus and {M}-polyfolds---an introduction},
    SERIES = {32$\sp {\rm o}$ Col\'oquio Brasileiro de Matem\'atica},
 PUBLISHER = {Instituto Nacional de Matem\'atica Pura e Aplicada (IMPA), Rio de Janeiro},
      YEAR = {2019},
     PAGES = {138}, }

@article{Hofer2009b,
    AUTHOR = {Hofer, Helmut and Wysocki, Krzysztof and Zehnder, Eduard},
     TITLE = {A general {F}redholm theory. {III}. {F}redholm functors and
              polyfolds},
   JOURNAL = {Geom. Topol.},
  FJOURNAL = {Geometry \& Topology},
    VOLUME = {13},
      YEAR = {2009},
    NUMBER = {4},
     PAGES = {2279--2387},  }

@article{Hofer2009a,
    AUTHOR = {Hofer, Helmut and Wysocki, Krzysztof and Zehnder, Eduard},
     TITLE = {A general {F}redholm theory. {II}. {I}mplicit function
              theorems},
   JOURNAL = {Geom. Funct. Anal.},
  FJOURNAL = {Geometric and Functional Analysis},
    VOLUME = {19},
      YEAR = {2009},
    NUMBER = {1},
     PAGES = {206--293},  }

@article{Hofer2007,
    AUTHOR = {Hofer, Helmut and Wysocki, Krzysztof and Zehnder, Eduard},
     TITLE = {A general {F}redholm theory. {I}. {A} splicing-based
              differential geometry},
   JOURNAL = {J. Eur. Math. Soc. (JEMS)},
  FJOURNAL = {Journal of the European Mathematical Society (JEMS)},
    VOLUME = {9},
      YEAR = {2007},
    NUMBER = {4},
     PAGES = {841--876},  }

@incollection{Hofer2021,
    AUTHOR = {Hofer, Helmut and Wysocki, Krzysztof and Zehnder, Eduard},
    TITLE = {Polyfold and {F}redholm theory},
    BOOKTITLE = {Ergebnisse der Mathematik und ihrer Grenzgebiete. 3. Folge. A Series of Modern Surveys in Mathematics},
    VOLUME = {72},
 PUBLISHER = {Springer, Cham},
      YEAR = {2021},
     PAGES = {xxii+1001},  }

@article{Andersdotter2024,
    author = {Emma Andersdotter and Daniel Persson and Fredrik Ohlsson},
    title = {Equivariant Manifold Neural {ODE}s and Differential Invariants},
    journal = {Journal of Machine Learning Research},
    volume = {26},
    pages = {1-33},
    year = {2025}
}

@article{Aahag2025,
  author        = {{\AA}hag, Per and  Czyż, Rafał  and Samuelsson Kalm, Håkan and Persson, Aron},
  title         = {On manifold-like polyfolds as differential geometrical objects with applications in complex geometry},
  year          = {2025},
 journal = {arxiv:2401.09875 [math.DG]},
}

@Book{Pata2019,
  author    = {Pata, Vittorino},
  publisher = {Springer International Publishing},
  title     = {Fixed Point Theorems and Applications},
  year      = {2019},
  journal   = {UNITEXT},  }

@Article{Bhat2000,
  author    = {Bhat, Sanjay P. and Bernstein, Dennis S.},
  journal   = {SIAM Journal on Control and Optimization},
  title     = {Finite-Time Stability of Continuous Autonomous Systems},
  year      = {2000},
  number    = {3},
  pages     = {751--766},
  volume    = {38},
  publisher = {Society for Industrial & Applied Mathematics (SIAM)},}

@inproceedings{Chen2018,
     author = {Chen, Ricky T. Q. and Rubanova, Yulia and Bettencourt, Jesse and Duvenaud, David K},
    booktitle = {Advances in Neural Information Processing Systems},
    pages = {},
    publisher = {Curran Associates, Inc.},
    title = {Neural Ordinary Differential Equations},
    volume = {31},
    year = {2018}, }

@inproceedings{Falorsi2020,
    author = {Luca Falorsi and Patrick Forré},
    title = {Neural Ordinary Differential Equations on Manifolds},
    booktitle = {Proceedings of the INNF+ Workshop of the International Conference on Machine Learning (ICML)},
    year = {2020}
}

@article{Haber2018,
    author = {Eldad Haber and Lars Ruthotto},
    title = {Stable architectures for deep neural networks},
    journal = {Inverse Problems},
    year = {2018},
    pages = {014004},
    volume = {34}
}

@inproceedings{Lipman2023,
    author = {Yaron Lipman and Ricky T.~Q.~Chen and Heli Ben-Hamu and Maximilian Nickel and Matt Le},
    title = {Flow Matching for Generative Modeling},
    booktitle = {Proceedings of the 11th International Conference on Learning Representations (ICLR)},
    year = {2023}
}

@inproceedings{Lou2020,
     author = {Lou, Aaron and Lim, Derek and Katsman, Isay and Huang, Leo and Jiang, Qingxuan and Lim, Ser Nam and De Sa, Christopher M},
    booktitle = {Advances in Neural Information Processing Systems},
    pages = {17548--17558},
    publisher = {Curran Associates, Inc.},
    title = {Neural Manifold Ordinary Differential Equations},
    volume = {33},
    year = {2020}, }

@inproceedings{Mathieu2020,
    author = {Mathieu, Emile and Nickel, Maximilian},
    booktitle = {Advances in Neural Information Processing Systems},
    pages = {2503--2515},
    publisher = {Curran Associates, Inc.},
    title = {Riemannian Continuous Normalizing Flows},
    volume = {33},
    year = {2020}, }

@article{Tong2024,
    author = {Alexander Tong and Kilian Fatras and Nikolay Malkin and Guillaume Huguet and Yanlei Zhang and Jarrid Rector-Brooks and Guy Wolf and Yoshua Bengio},
    title = {Improving and Generalizing Flow-Based Generative Models with Minibatch Optimal Transport},
    journal = {Transactions of Machine Learning Research},
    year = {2024},
    volume = {03}
}

@inproceedings{Zhang2020,
    title={Approximation Capabilities of Neural {ODE}s and Invertible Residual Networks}, 
    author={Han Zhang and Xi Gao and Jacob Unterman and Tom Arodz},
    year={2020},
    booktitle = {Proceedings of the 37th International Conference on Machine Learning},
    publisher = {PMLR},
    volume = {119},
    pages = {11086-11095}
}

@phdthesis{LeCun1987,
    author = {Yan LeCun},
    title = {Modèles connexionistes de l’apprentissage},
    school = {Université de Paris VI},
    year = {1987}
}

@article{Bourlard1988,
    author = {Bourlard, H. and Kamp, Y.},
    volume = {59},
    pages = {291-294},
    title = {Auto-association by multilayer perceptrons and singular value decomposition},
    journal = {Biological Cybernetics},
    year = {1988}
}

@inproceedings{Albergo2023,
    author = {Michael S. Albergo and Eric Vanden-Eijnden},
    title = {Building Normalizing Flows with Stochastic Interpolants},
    booktitle = {Proceedings of the International Conference on Learning Representations (ICLR)},
    year = {2023}
}

@article{Thorpe2023,
    author = {Matthew Thorpe and Yves van Gennip},
    title = {Deep limits of residual neural networks},
    journal = {Research in the Mathematical Sciences},
    year = {2023},
    volume = {10},
    number = {6}
}

@article{RuizBalet2023,
    author = {Dom\`{e}nec Ruiz-Balet and Enrique Zuazua},
    title = {Neural {ODE} Control for Classification, Approximation, and Transport},
    journal = {SIAM Review},
    volume = {65},
    pages = {735-773},
    year = {2023}
}

@inproceedings{Liu2023,
    author = {Xingchao Liu and Chengyue Gong and Qiang Liu},
    title = {Flow straight and fast: Learning to generate and transfer data with rectified flow},
    booktitle = {Proceedings of the 11th International Conference on Learning Representations (ICLR)},
    year = {2023}
}

@article{Baldi1989,
    author = {Baldi, P. and Hornik, K.},
    title = {Neural networks and principal components analysis: Learning from examples without local minima},
    journal = {Neural Networks},
    volume = {2},
    pages = {53-58},
    year = {1989}
}

@article{Hinton1989,
    author = {Hinton, G. E.},
    title = {Connectionist learning procedures},
    journal = {Artificial Intelligence},
    volume = {40},
    pages = {185-234},
    year = {1989}
}

@Article{Fabert2016,
  author    = {Fabert, Oliver and Fish, Joel W. and Golovko, Roman and Wehrheim, Katrin},
  journal   = {EMS Surveys in Mathematical Sciences},
  title     = {Polyfolds: A first and second look},
  year      = {2016},
  number    = {2},
  pages     = {131--208},
  volume    = {3}
}

@misc{torchdiffeq,
	author={Chen, Ricky T. Q.},
	title={torchdiffeq},
	year={2018},
	url={https://github.com/rtqichen/torchdiffeq},
}

@Article{Memoli2005,
  author    = {Mémoli, Facundo and Sapiro, Guillermo},
  journal   = {SIAM Journal on Applied Mathematics},
  title     = {Distance Functions and Geodesics on Submanifolds of $\mathbb{R}^d$ and Point Clouds},
  year      = {2005},
  number    = {4},
  pages     = {1227--1260},
  volume    = {65},
  publisher = {Society for Industrial & Applied Mathematics (SIAM)},  }

@inproceedings{Glorot2010,
    author = {Glorot, Xavier and Bengio, Yoshua},
    title = {Understanding the difficulty of training deep feedforward neural networks},
    booktitle = {Proceedings of the Thirteenth International Conference on Artificial Intelligence and Statistics},
    pages = 	 {249--256},
    year = 	 {2010},
    volume = 	 {9},
    publisher =    {PMLR} }

@misc{PolynodeCode2026,
	author={Friedrich, Alexander and Vigren Näslund, Viktor},
	title={polynode},
	year={2026},
	url={https://github.com/turbotage/PolyNODE},
}

\newpage
\appendix
\onecolumn

\section{Experimental Details}
\label{app:ex_details}
The implementation for pytorch for the experiments below can be found in \cite{PolynodeCode2026} at \texttt{https://github.com/turbotage/PolyNODE}.
\subsection{Reconstruction Experiments}
\label{app:ex_reconst}
In all reconstruction experiments we employ a reduced setup for the vector fields detailed in Table~\ref{table:vectorfield_reduced}. Recall that we keep the speed of $Y_1$ constant to ensure a total time $T=1$ and a latent time of $1/2$. The rates of the compressing vector fields are constant as well, $k_i = 25$, and $a=1/2$. 

\begin{table}[ht!]
\centering
\caption{Structure of vector fields for reconstruction experiments}
\label{table:vectorfield_reduced}
\begin{tabular}{lp{4.5cm}p{2.6cm}l}
\hline
\textbf{Region} & \textbf{Shape of Vector Field} & \textbf{Neural Networks} & \textbf{Input} \\
\hline
$\{\tau < \tau_0\}$ & $(C, Y_2, ..., Y_{n+m+1})$ & $\{Y\}_2^{n+m+1}$ & $(\tau, x,y)$ \\
\hline
$\{\tau_0 < \tau < \tau_1\}$ & $(C, 0,...,0 , Y_{n+2}, ... Y_{n+m+1})$,  $\{Y_i\}_{i=n+2}^{n+m+1}$ as in~\eqref{eq:comp_vf},  & - & $(\tau, x,y)$ \\
\hline
$\{\tau_1 < \tau < \tau_2\}$ & $(C, 0,..., 0)$ & - & - \\
\hline
 $ \{\tau > \tau_2\}$ & $(C, Y_2, ..., Y_{n+m+1})$ & $\{Y\}_2^{n+m+1}$ & $(\tau, x,y)$ \\
\hline
\end{tabular}
\end{table}

All neural networks are sequential with four fully connected layers of width $200$ and $\tanh$ as activation function. The first three layers have bias, the last one does not. The bias are initialised as $0$ and  linear layers are initialised according to a normal distribution.

Using the Euler ODE solver implemented in the {\tt torchdiffeq} python package with $500$ time steps, along with the adjoint method for the backpropagation of the same package, \cite{torchdiffeq}, we solve the flow equation in two stages, each integrating for half the total time. Thus, after the first stage, the flow maps the input data into the latent space. There, we project to the latent space to eliminate the remaining numerical error in the latent components, which we find to be of order $10^{-8}$, then we continue the flow. The backwards pass for the compressing vector fields is customised as described in Appendix~\ref{app:backprop}.
The models are trained using RMSprop, with momentum $0.3$.

The experiments were run on a cluster, using the CUDA backend of PyTorch with NVIDIA Tesla T4 GPUs, $16$GB VRAM, and  Intel(R) Xeon(R) Gold 6226R CPUs ($2.90$GHz).

\paragraph{Spiral}
\label{app:ex_spiral}
For the spiral experiment, consider the following parametrisation where $v\in \R$, $e_2=(0,1,0,0)$. 
\begin{align*}
    f: [0,2\pi] \to \R^3\,,\quad s \mapsto (1 + 0.5 s) (\cos(v s ), \sin(v s ) , 0) + 6 e_2.
\end{align*}
We vary the number of turns from $N=0.5$ to $N=5$ by varying the speed $v$. 
The input training data is equidistantly sampled according to intrinsic distance $d_S$ of the spiral, which can be computed analytically. The number of sample points is 20 times the length of the spiral, rounded down, and capped at $5000$ due to memory constraints during training. For the validation data $0.3$ times the number of training samples are drawn randomly in the angle domain $(0, 2\pi)$.

We choose $\Omega^2_1$ with $\tau_1=0$ and $\tau_2=1$ as an M-polyfold, and $\tau_0 =-3$ for the onset of the compressing vector fields. For the embedding we choose $\tau_\mathcal{X}=-7$ and $\tau_\mathcal{Y}=8$, hence the speed is $C=15$. The models are trained for about $27000$ epochs unless the loss does not improve significantly. This happened only for $N=0.5$ and $N=1$, which stopped at about $8000$ and $18000$ epochs respectively.

\begin{figure}[ht]
    \centering
    \begin{subfigure}[b]{0.245\textwidth}
        \includegraphics[width=\textwidth]{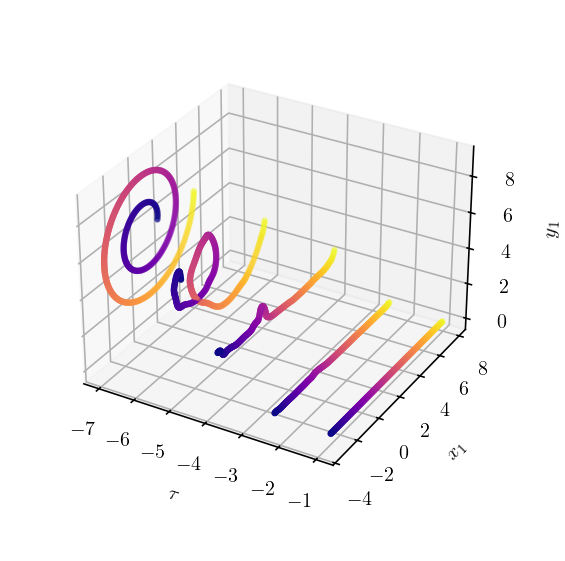}
    \end{subfigure}
    \begin{subfigure}[b]{0.245\textwidth}
        \includegraphics[width=\textwidth]{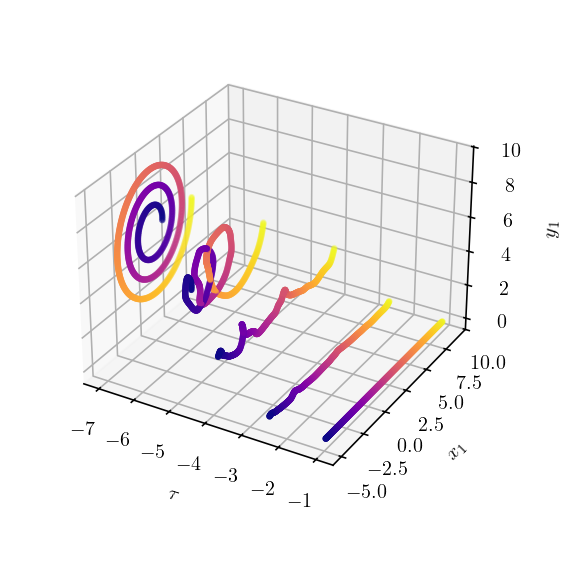}
    \end{subfigure}    
    \begin{subfigure}[b]{0.245\textwidth}
        \includegraphics[width=\textwidth]{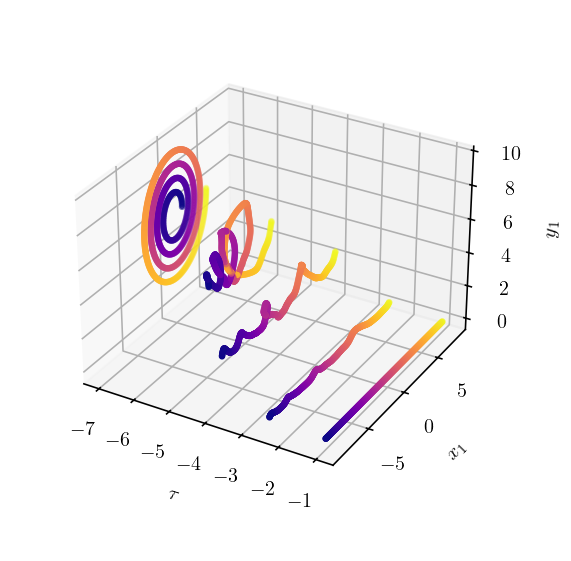}
    \end{subfigure}
    \begin{subfigure}[b]{0.245\textwidth}
        \includegraphics[width=\textwidth]{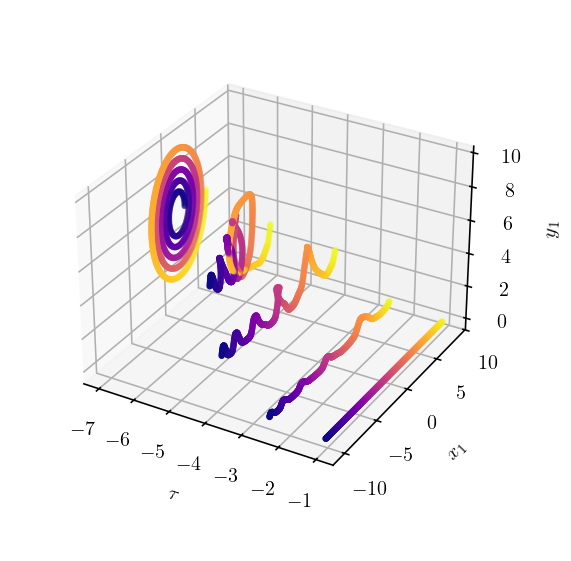}
    \end{subfigure}
    \caption{Unwinding and encoding of spirals with number of turns ranging from $N=2$ to $N=5$}\label{fig:unwinding_appendix}
\end{figure}

\begin{figure}[ht!]
    \centering
    \begin{subfigure}[b]{0.245\textwidth}
        \includegraphics[width=\textwidth]{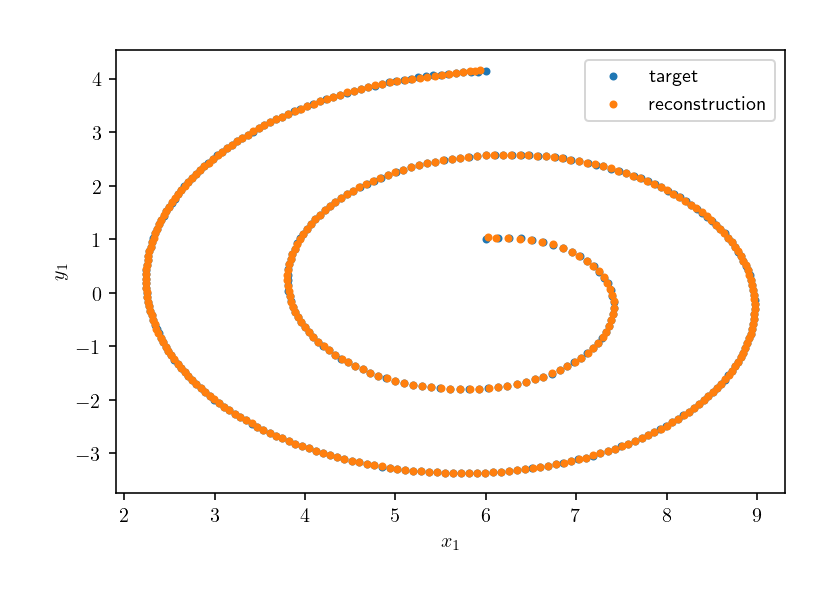}
    \end{subfigure}
    \begin{subfigure}[b]{0.245\textwidth}
        \includegraphics[width=\textwidth]{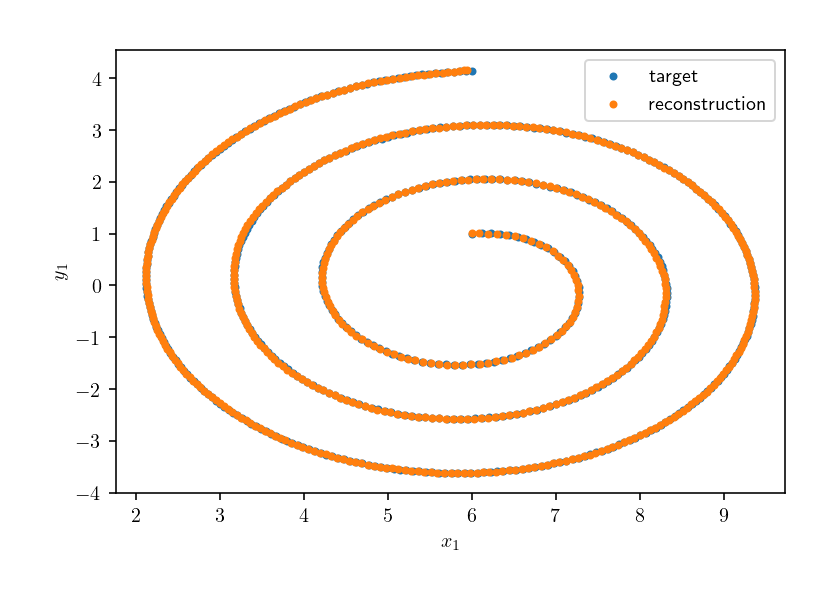}
    \end{subfigure}
    \begin{subfigure}[b]{0.245\textwidth}
        \includegraphics[width=\textwidth]{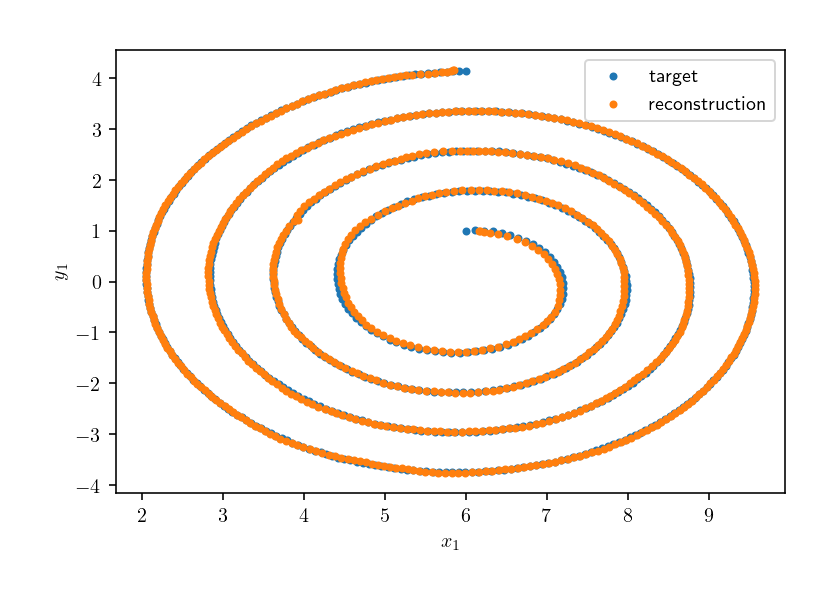}
    \end{subfigure}
    \begin{subfigure}[b]{0.245\textwidth}
        \includegraphics[width=\textwidth]{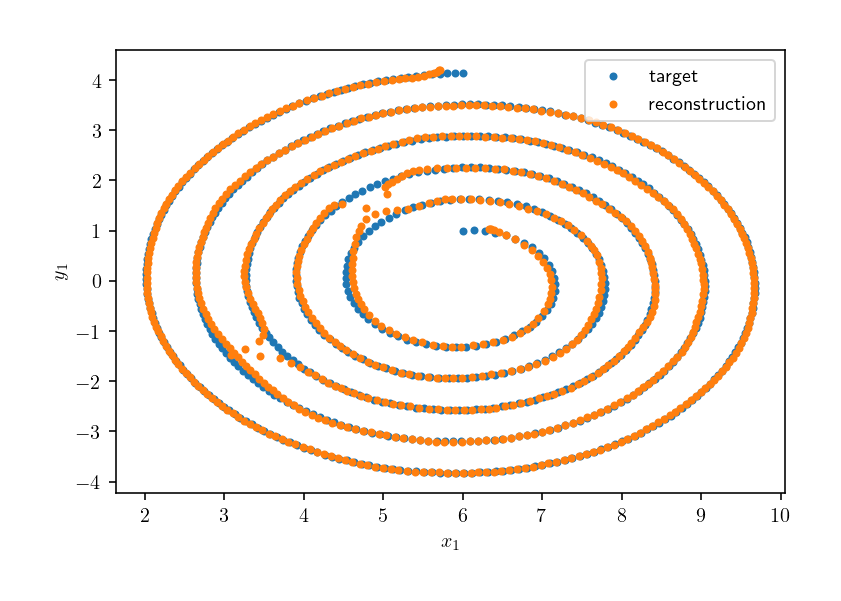}
    \end{subfigure}

    \caption{Reconstruction (2d projection) of spirals with number of turns ranging from $N=2$ to $N=5$}\label{fig:reconstruction_sprial_appendix}
\end{figure}

\begin{figure}[ht!]
    \centering
        \includegraphics[width= 0.54\textwidth]{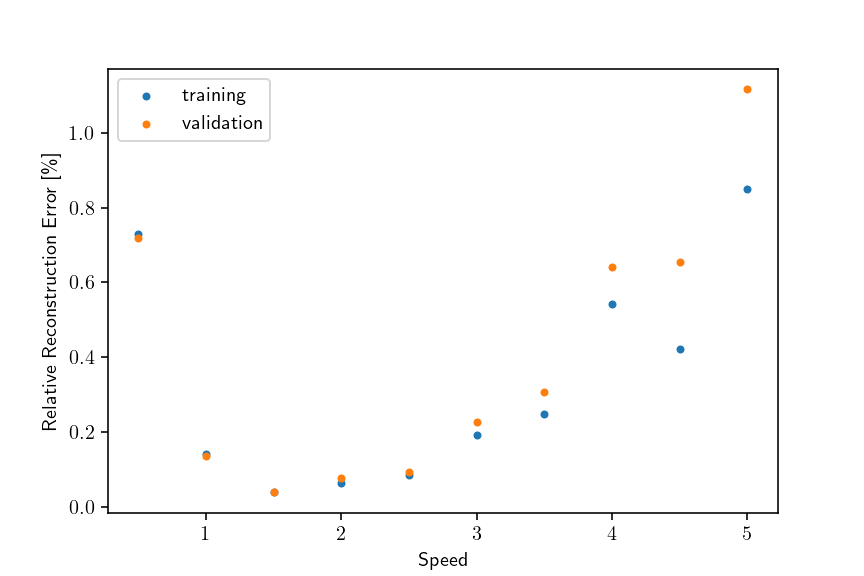}
        \caption{Mean euclidean reconstruction error relative to extrinsic spiral diameter}
        \label{fig:recon_error}
\end{figure}

Figure~\ref{fig:unwinding_appendix} shows time slices of the unwinding and encoding of spirals with $N=2$ to $N=5$ and Figure~\ref{fig:reconstruction_sprial_appendix} shows the corresponding projected reconstructions. Up to $N=3$ the encoding and reconstruction is excellent, but these plots also illustrate that our autoencoding NODE achieves very good encoding, even if the reconstruction suffers slightly. This observation is supported by the mean reconstruction error plot in Figure~\ref{fig:recon_error} and the fact the monotonicity error is below $0.02\%$ in all experiments.
This is likely due to the rather strong approximate isometry loss $L_3$ for the unwinding and the explicit compressing vector field. Hence the encoder and decoder are not symmetric.

\paragraph{Sphere}
\label{app:ex_sphere}
Consider a two dimensional round sphere of radius $1$ in $\R^4$ around the point $x_0 = (0,0,0,3)$ given by $\{ (z_1,  z_2, 0,  z_3) \mid (z_1,z_2,z_3) \in S^2_1(x_0) \subset \R^3 \}$. The $z\in S^2_1$ in turn are parametrised by spherical coordinates. The  training data is sampled from the sphere, equidistantly on a grid  in the angle domain $(0, 2\pi)\times (-\pi, \pi)$ with $6000$ points in total. The $1800$  validation data points are sampled randomly from the angle domain.
We choose $\Omega^1_2$ with $\tau_1=0$ and $\tau_2=3$ as an M-polyfold with $\tau_0 =-3$ for the onset of the compressing vector fields. For the embeddings of the samples  $ \mathcal{X}= \{p_i\}_{i\in I} = \{z_i+ \tau_\mathcal{X} e_1 \}_{i\in I} $ and $ \mathcal{Y} = \{z_i+ \tau_\mathcal{Y} e_1 \}_{i\in I} $ we choose $\tau_\mathcal{X} = -7$ as well as $\tau_\mathcal{Y}=10$. This leaves us with a speed of $C=17$. The model was trained for about $10000$ epochs.

\subsection{Classification Experiments}
\label{app:ex_classification}
Both the radial and angular classification experiments use the same neural network architecture. The layers consist of a linear input layer with input size $4$ (spatial + time) and output size $512$, followed by two hidden layers with input and output size $512$. A skip connection is used between the hidden layers. The final output layer is a linear layer with input size $512$ and output size $3$. All layers are initialised using Xavier uniform initialisation \cite{Glorot2010}, and the hidden-layer biases are zero-initialised. ReLU is used as the nonlinearity between layers. The models are trained using the Adam optimiser. A reduce-on-plateau learning rate scheduler is used with a patience of $20$ epochs and a factor of $0.8$. Both experiments run for $300$ epochs, and the batch size is $32$.

The target points are placed equidistantly to the latent line at the points $(-3,4,0)$ for $l_1$, $(0,5,0)$ for $l_2$, and $(3,4,0)$ for $l_3$. For the attractor terms in \eqref{eq:classification_vf} we use $C = 50$ and $k=64/d_{\min}$, where $d_{\min}$ is the minimum distance between any two target points ($\sqrt{2}$ in this setup).

Experiments were conducted on the CUDA backend of PyTorch with an NVIDIA RTX 4090 GPU, with CUDNN set to deterministic mode and a fixed random seed of $42$, and Intel(R) Core i9-14900K CPU.

\section{Changes to the Backpropagation}
\label{app:backprop}

The derivative of the compressing vector field \eqref{eq:comp_vf} has a singularity of the form $|y_i|^{a -1}$ at $y_i=0$. However, we need to calculate the derivative during training with gradient descent methods. 
Thus, we introduce a cut-off for the vector field for small values of $y_i$, and implement a corresponding derivative which is used in the custom backward method of the vector field.
For simplicity we use a cut-off function $\varphi$ based on a polynomial,
\begin{align*}
    \varphi(x) = 
    \begin{cases}
       0 & \text{for } x<0 \\
        (3-2x)x^2 & \text{for }  0\leq x\leq 1 \\
         1 & \text{for } x>1 \\
    \end{cases}\,,
\end{align*}
which has derivative
\begin{align*}
    \varphi'(x) = 
    \begin{cases}
       0 & \text{for } x<0 \\
        6(1-x)x & \text{for } 0\leq x\leq 1 \\
     0 & \text{for } x>1 \\
    \end{cases} \,.
\end{align*}
Scaling $\varphi$ gives a cut-off function $\varphi_{b,c}:=\varphi((x-b)/(c-b))$ in the interval $(b,c)$, $b\geq 0$.
In the forward pass the vector field then reads 
\begin{align*}
    Y_{1+n+j}(p) &= -k_j(p) \sign(y_j) |y_j|^a \varphi_{b,c}(|y_j|)\,.
\end{align*}
In the backwards pass the derivative is given as
\begin{align*}
  \nabla  Y_{1+n+j}(p) &= -\nabla k_j(p) \sign(y_j) |y_j|^a \varphi_{b,c}(|y_j|) \\
  & \quad -k_j(p) a|y_j|^{a-1} \varphi_{b,c}(|y_j|) e_j \\
  & \quad -k_j(p) |y_j|^a \frac{\varphi_{b,c}'(|y_j|)}{c-b} e_j \,,
\end{align*}
which does not have a singularity at $0$.

In the experiments of Section~\ref{subsec:reconstruction} we use $b=10^{-7}$, and $c=10^{-6}$. This introduces numerical errors in the latent state in the compressed directions $y$ which are at most of the order $b$. In these experiments we find errors of the order $10^{-8}$ and use a projection to the latent plane to eliminate them as described in Appendix~\ref{app:ex_details}.

\end{document}